\documentclass[lettersize,twoside,journal]{IEEEtran}
\usepackage{amsmath,amsfonts}
\usepackage{algorithmic}
\usepackage{algorithm}
\usepackage{array}
\usepackage[caption=false,font=normalsize,labelfont=sf,textfont=sf]{subfig}
\usepackage{textcomp}
\usepackage{stfloats}
\usepackage{url}
\usepackage{verbatim}
\usepackage{graphicx}
\usepackage{cite}
\usepackage{mathrsfs}
\usepackage{color}
\usepackage{colortbl}
\definecolor{Ocean}{RGB}{129,154,254}
\definecolor{orange}{RGB}{254,144,95}
\usepackage{bbding}
\usepackage{pifont}
\usepackage{tabularx}
\usepackage{multirow}
\usepackage{diagbox}
\usepackage{xcolor}

\usepackage{booktabs}
\usepackage{makecell}
\usepackage{threeparttable}

\begin{document}

\title{SphereSOD: Geometry-Structure Coupled Learning for 360 Salient Object Detection}

\author{ Junsong~Zhang$^\dagger$, 
Zhijie~Shen$^\dagger$, Shuai~Zheng, Feng~Li, Runmin~Cong,  Yao~Zhao,, Chunyu~Lin*,

\thanks{}
\thanks{}}

\markboth{}%
{ZHANG \MakeLowercase{\textit{et al.}}: 
SphereSOD: Geometry-Structure Coupled Learning for 360 Salient Object Detection}


\maketitle

\begin{abstract}

360° salient object detection (SOD) aims to accurately segment salient regions across a full field of view. However, equirectangular projection (ERP) introduces severe spatial distortion when mapping the spherical domain onto a planar representation. Existing methods mainly focus on compensating projection distortion while overlooking the interaction between panoramic geometry and salient object structure during feature perception and prediction refinement. To this end, we propose SphereSOD, an ERP-native framework that couples panoramic geometry with evolving salient structures. Specifically, spherical geometry governs feature sampling and spatial weighting, while coarse-grained saliency and contour prediction influence context aggregation during the progressive decoding process. SphereSOD first initializes deformable sampling based on spherical projection geometry and then employs bounded, content-adaptive offsets, yielding features that are better aligned with the underlying panoramic geometry. Subsequently, the decoder performs structure-guided context aggregation and progressive refinement to recover complete salient regions and accurate boundaries. 
Extensive experiments on three public 360° SOD benchmarks demonstrate state-of-the-art performance and a favorable accuracy-efficiency trade-off, supporting structure-preserving inference directly in ERP space as a promising alternative to projection-heavy panoramic pipelines.  

\end{abstract}


\section{Introduction}
\begin{figure}[t]
\centering
\includegraphics[scale=0.38]{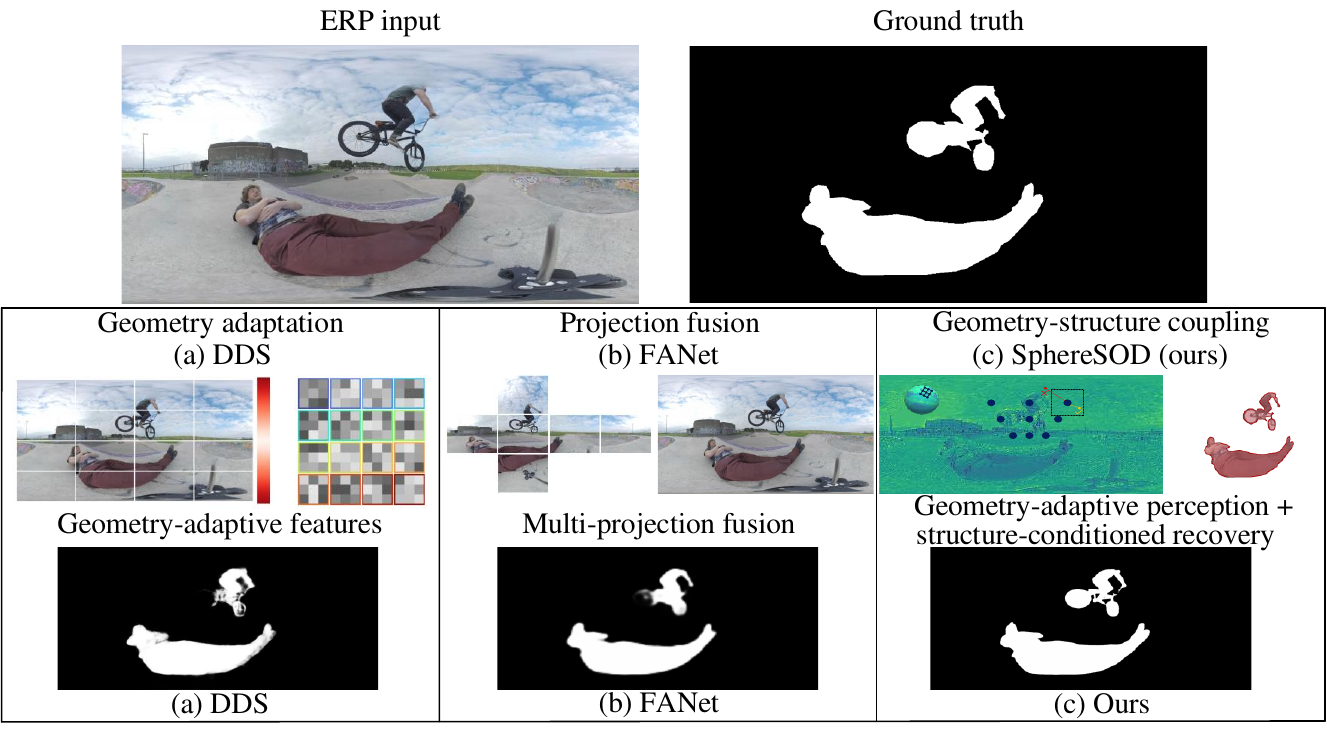}
\caption{\textbf {Brief comparison between SphereSOD and other 360$^{\circ}$ SOD methods. (a) DDS~\cite{r20} splits ERP images into regions with region-specific convolution kernels. (b) FANet~\cite{r22} adaptively fuses the ERP image with six cube-map faces. (c) Our SphereSOD instead couples geometry-adaptive perception with structure-conditioned recovery directly in ERP space.
}}
\label{fig1}
\end{figure}

With the rapid development of augmented reality (AR), virtual reality (VR), and immersive visual applications, panoramic imaging has become an important representation for capturing surrounding environments. Unlike conventional perspective images, 360$^{\circ}$ images provide a complete field of view within a single observation, offering rich contextual information for comprehensive scene understanding. Salient object detection (SOD), which aims to identify and segment visually important regions, provides an effective means of extracting informative content from such large-field-of-view imagery and benefits various downstream tasks, including semantic segmentation \cite{r58,r59}, scene classification \cite{r61}, image editing \cite{r66,r67,r68}, and object tracking \cite{r70,r71}.


Although perspective-image SOD methods have achieved remarkable progress \cite{r10,r56,r72,r73,r74,r76,r11,r81,r82,r83,r85}, directly applying them to 360$^{\circ}$ images remains challenging. The core difficulty arises from the mismatch between spherical scene geometry and planar image operations. Equirectangular projection (ERP) introduces geometric distortion and places horizontally adjacent spherical locations at opposite image boundaries. These geometric inconsistencies alter the spatial relationships assumed by conventional feature sampling and context aggregation, causing errors in feature perception to propagate into incomplete salient regions and inaccurate object boundaries. Therefore, the challenge of 360$^{\circ}$ SOD is not merely to compensate for projection distortion, but to keep panoramic geometry involved in the recovery of salient object structure.


Existing 360$^{\circ}$ SOD methods address ERP distortion through two common, sometimes overlapping directions. One direction adapts feature extraction directly to non-uniform ERP geometry. For example, DDS \cite{r20} learns region-specific convolution kernels for partitioned ERP regions, while DATFormer \cite{r25} introduces distortion-adaptive modules and relation matrices into a transformer. These methods improve feature representation under spatially varying distortion.
However, geometric adaptation is typically implemented as a representation-level operation, and how panoramic geometry should interact with evolving region and boundary estimates during subsequent prediction refinement remains less explored.


Another direction introduces alternative projections for less-distorted local observations. For example, \cite{r21} combines perspective views with object-level semantic ranking, FANet \cite{r22} selectively fuses ERP and cubemap projection (CMP) features, MPFRNet \cite{r23} jointly exploits ERP and four cube-unfolding (CU) views, and SIHENet and SCFANet \cite{r42,r41} combine local CMP cues with global ERP information. Although complementary views improve local perception, projection conversion divides the spherical domain into separate representations and requires extra transformation, alignment, and fusion, which may break the continuity of boundary-spanning objects and add processing overhead. More fundamentally, reducing distortion through representation conversion does not by itself determine how spherical geometry should guide the recovery of complete salient regions and boundaries.


These observations suggest that panoramic geometry modeling and salient-structure recovery should not be treated as isolated processes. We therefore propose SphereSOD, an ERP-native framework that couples panoramic geometry with evolving salient structures across feature perception and progressive decoding. Here, salient object structure refers to the spatial organization of foreground regions and their boundaries, represented by evolving saliency and contour predictions during decoding. In SphereSOD, spherical geometry governs where visual evidence is sampled and how spatial context is weighted, whereas saliency and contour predictions determine which regions and boundaries to emphasize during aggregation. Their interaction connects geometry-adaptive feature perception with structure-conditioned progressive recovery directly in ERP space, without additional projection conversion.

Specifically, at the encoder stage, we propose the Prior-Guided Multi-scale Encoder (PG-ME). PG-ME initializes deformable sampling from spherical projection geometry, enabling feature extraction to adapt to latitude-dependent distortion in ERP images. It further introduces bounded content-adaptive offsets to constrain sampling within relevant local regions and reduce interference from irrelevant areas. Meanwhile, circular continuity preserves the horizontal consistency of ERP. In this way, PG-ME produces geometry-aligned features while maintaining local salient information.

At the decoder stage, we design the Dual-branch Prior-guided Progressive Decoder (DPPD) for structure-guided context aggregation and progressive refinement. Guided by task-related structural cues, DPPD aggregates saliency-relevant context and progressively reconstructs complete salient regions and accurate boundaries. Furthermore, overlap-normalized upsampling and probability-context residual refinement preserve structural details during resolution recovery. As shown in Fig.~\ref{fig1}, SphereSOD yields more complete object structures and clearer boundaries than existing 360$^{\circ}$ SOD methods.

We conduct extensive experiments on three datasets, including 360-SOD \cite{r20}, 360-SSOD \cite{r21}, and ODI-SOD \cite{r24}. Results show that SphereSOD achieves state-of-the-art performance with a favorable accuracy-efficiency trade-off. The main contributions are summarized as follows:

\begin{itemize}

\item We propose SphereSOD, an ERP-native framework that couples panoramic geometry with evolving salient structures, connecting geometry-adaptive feature perception to structure-conditioned progressive recovery for 360$^{\circ}$ SOD.

\item We develop PG-ME and DPPD to realize geometry-adaptive feature perception and structure-aware progressive refinement, respectively, enabling accurate salient object recovery without multi-projection fusion.

\item Extensive experiments on three public benchmarks show that SphereSOD achieves state-of-the-art performance with favorable computational efficiency.

\end{itemize}

\section{Related Work}
\subsection{SOD in 2D Images}
SOD in 2D images has been extensively studied, with methods broadly divided into traditional handcrafted-feature and deep learning-based approaches. Early methods rely on manually designed cues such as texture \cite{r1,r2}, color \cite{r3,r4}, and background priors \cite{r5,r6}, but their dependence on domain knowledge limits adaptability and efficiency.

With the development of deep learning, CNN-based methods have become dominant. \cite{r7} integrates global and local information for saliency modeling, \cite{r8} enhances multi-scale context through a cascaded structure, \cite{r87} refines features and models cross-modal interactions, and \cite{r9} fuses high-level semantics with low-level details via short connections and multi-scale supervision. Moreover, \cite{r10,r11,r12} introduce edge maps as auxiliary guidance for more accurate localization.

More recently, Transformer-based methods model long-range dependencies beyond CNN receptive fields. \cite{r13} proposes a pure Transformer for sequence-to-sequence saliency prediction with saliency and edge supervision, \cite{r14} combines CNNs and Transformers for complementary local-global modeling, and \cite{r15} adopts a dual-branch structure to separately learn local details and global context.

\subsection{SOD in 360$^{\circ }$ Omnidirectional Images}
With the rapid development of virtual reality (VR), saliency detection in panoramic images has attracted increasing attention. Early studies mainly focus on fixation prediction in 360$^{\circ}$ images and videos. \cite{r16} employs generative adversarial imitation learning (GAIL) to simulate human head-movement trajectories. \cite{r17} proposes a spatiotemporal network for 360$^{\circ}$ video saliency prediction with cube filling to reduce distortion. Salnet360 \cite{r18} adapts conventional 2D SOD models to panoramic scenes by partitioning each image into six regions and merging CNN-based saliency maps. \cite{r19} introduces a spherical U-Net with shared spherical convolution kernels. However, fixation prediction models human visual attention, while 360$^{\circ}$ SOD requires pixel-level localization with accurate boundaries, making it more challenging and less explored.

For 360$^{\circ}$ SOD, \cite{r20} builds the first dedicated dataset with a distortion-adaptive module and multi-scale context integration, and \cite{r24} contributes a large-scale dataset with a sample-adaptive view transformer for distortion, boundary discontinuity, and scale variation. \cite{r21} formulates SOD as a multi-stage task with object-level semantic saliency ranking and leverages 2D images to improve accuracy. DATFormer \cite{r25} integrates distortion-adaptive modules with relation matrices. 
FANet \cite{r22} adaptively fuses ERP images and six cube-map faces. MPFRNet \cite{r23} dynamically weights multiple cube-map projections to alleviate incomplete object representation. SCFANet \cite{r42} captures global ERP cues with ViT and local cube-map cues via CNN, while SIHENet \cite{r41} embeds ERP features into less-distorted CMP contexts to exploit cross-projection and cross-level complementarity. \cite{r88} learns saliency features from tangent-projection images and models their geometric correlations with ERP images.

Although these methods improve 360$^{\circ}$ SOD performance, they still exhibit notable limitations. ERP distortion-adaptive methods mainly focus on feature extraction, with limited consideration of panoramic geometry in subsequent structural recovery. Multi-projection methods alleviate local distortion but introduce extra projection and fusion overhead and may disrupt the continuity of boundary-spanning objects.
To address these issues, we propose SphereSOD, an ERP-native framework that connects geometry-adaptive perception with structure-conditioned recovery by jointly modeling panoramic geometry and evolving salient structures throughout encoding and decoding. Specifically, we design a Prior-Guided Multi-scale Encoder (PG-ME) to adapt feature sampling to spherical projection geometry and a Dual-branch Prior-guided Progressive Decoder (DPPD) to aggregate context and progressively refine predictions under evolving saliency and contour cues. This design recovers complete salient regions and clear boundaries directly in ERP space, without multi-projection fusion.


\begin{figure*}[t]
\centering
\includegraphics[scale=1]{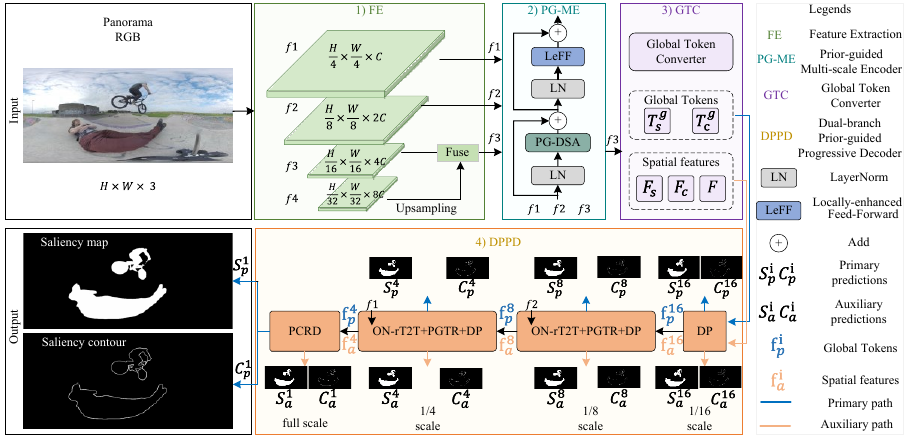}
\caption{\textbf {Overview of our proposed SphereSOD. The SphereSOD comprises four major parts: 
feature extraction based on Swin-s,
prior-guided multi-scale encoder (PG-ME), 
global token converter (GTC), and the dual-branch prior-guided progressive decoder (DPPD). 
}}

\label{fig2}
\end{figure*}

\section{Method}
\subsection{Architecture Overview}
In Fig. \ref{fig2}, we show our complete framework. The overall design of SphereSOD follows a geometry-structure coupled paradigm: panoramic geometry guides where features are perceived, while evolving salient structures condition how predictions are progressively recovered, all directly in ERP space. SphereSOD takes a panorama as input and outputs a predicted saliency map and contour. 
During the encoder stage, we employ Swin-s to extract features at four different resolutions. To fully leverage the global semantics from the deepest layer of the Swin Transformer, the high-level semantic features $f4$ are concatenated with $f3$ along the channel dimension and then fused through an MLP, thereby yielding features that integrate both global semantics and local details. Then, these features are provided to the prior-guided multi-scale encoder (PG-ME), which applies the prior-guided deformable self-attention (PG-DSA) across three scales, leveraging spherical sampling prior, local semantic prior, and horizontal continuity prior to produce geometry-aligned features. Subsequently, $f3$ is passed through the global token converter (GTC), which transforms it into global saliency and contour contextual tokens and spatial features for the subsequent decoder. 
Finally, the dual-branch prior-guided progressive decoder (DPPD) is designed to progressively refine saliency and contour predictions from coarse scale to full resolution through overlap-normalized reverse T2T upsampling (ON-rT2T), prior-guided token refinement (PGTR), 
primary auxiliary dual prediction (DP), and probability-context residual decoder (PCRD). DPPD jointly optimizes a token-similarity-based primary branch for global semantic alignment and a projection-based auxiliary branch for local detail preservation. In this way, spherical geometry governs feature sampling and spatial weighting across PG-ME and DPPD, while evolving saliency and contour predictions condition progressive recovery, coupling panoramic geometry with salient structures throughout perception and decoding.

\subsection{Prior-guided Multi-scale Encoder (PG-ME)}
Panoramic SOD aims to localize the complete regions and precise boundaries of salient objects, which requires both local contextual discrimination and adaptation to the spherical distortion and horizontal continuity introduced by equirectangular projection (ERP). 
To this end, we introduce the PG-ME, which leverages the local semantic prior of salient objects to constrain deformable sampling to semantically relevant local neighborhoods via bounded content-adaptive offsets. Meanwhile, it incorporates the spherical sampling prior and circular padding to adapt to the distortion and continuity.

Specifically, as shown in Fig. \ref{fig2}, PG-ME is composed of two stacked standard transformer blocks at each scale. The input features are first normalized by LayerNorm and then fed into the core prior-guided deformable self-attention (PG-DSA) module, followed by a residual connection. Subsequently, the features pass through another LayerNorm and the Locally-enhanced Feed-Forward (LeFF) module, with a second residual connection applied to yield the final output of the PG-ME.

\begin{figure}[t]
\centering
\includegraphics[scale=0.78]{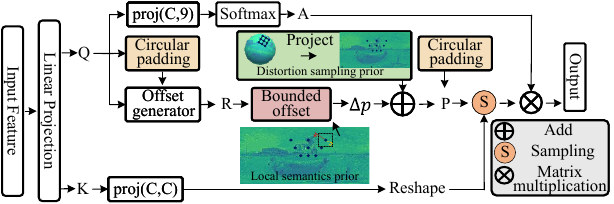}
\caption{\textbf {Overview of PG-DSA, which is supported by three priors.}}
\label{fig3}
\end{figure}

As illustrated in Fig. \ref{fig3}, in PG-DSA, the input features are first linearly projected into query $Q$ and key $K$. The query $Q$ is passed through a linear projection layer $proj$ to compute the attention scores of the sampling points, which are then normalized by Softmax to obtain the attention weights $A$. In addition, $Q$ is fed into the offset generator to predict the raw offsets $R$. To ensure the horizontal continuity of offset prediction in ERP, circular padding is applied in the offset generator. To focus on semantically relevant local contexts and prevent irrelevant long-range interference, we impose a local-semantics prior on the offset prediction, constraining the raw offsets $R$ to obtain the bounded residual offsets $\bigtriangleup p$:
\begin{equation}
\bigtriangleup p=tanh(R)\cdot  Rmax\cdot  g,  \quad g=\sigma (\gamma )
\end{equation}
where the $tanh$ function constrains the raw offsets to the range of $(-1, 1)$, and $Rmax$ denotes the maximum offset magnitude (set to 1.5 empirically). The gate $g$ serves as a learnable offset gate ($\sigma$ denotes the sigmoid function and $\gamma$ is a learnable parameter) to adaptively adjust the offset magnitude. 
The gate $g$ is initialized to 0.1 to suppress excessive offset magnitudes at the early training stage. Subsequently, the distortion sampling prior provides initial locations on the ERP plane. The bounded offset $\bigtriangleup p$ is then applied to these prior positions to derive the final sampling points $P$.
After the addition, circular padding is further applied to handle longitude continuity. Finally, the key $K$ is linearly projected and reshaped into multi-head features, bilinearly sampled at the sampled points, weighted by attention weights $A$, and projected to yield the final PG-DSA output.


\begin{figure}[t]
\centering
\includegraphics[scale=0.8]{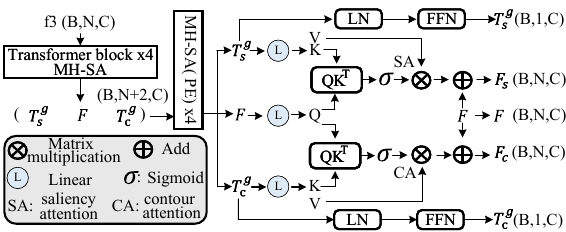}
\caption{\textbf {Architecture of the GTC. It generates global task tokens ($T_{s}^{g}$, $T_{c}^{g}$) for the primary path and task-enhanced spatial features ($F_{s}$, $F_{c}$, $F$) for the auxiliary path.}}
\label{fig4}
\end{figure}

\subsection{Global Token Converter (GTC)}
Although the features processed by PG-ME acquire local contextual awareness and panoramic geometric adaptability, they still lack global semantics for saliency and contour prediction. To model task-relevant contexts globally, we follow VST++ \cite{r89} and introduce the Global Token Converter (GTC). It is worth noting that the GTC is not the core contribution of this work; however, we describe it here for completeness.

As shown in Fig. \ref{fig4}, considering both high-level semantics and computational efficiency, the GTC is applied only to the 1/16 scale features. First, the input features $f3 \in R^{B\times  N\times  C}$  pass through four standard multi-head self-attention layers to establish global context. Then, we introduce two learnable global tokens, namely the saliency token $T_{s}^{g}$ and the contour token $T_{c}^{g}$, and concatenate them to the two ends of the patch token $F$, resulting in a sequence of length $N+2$. To further model the spatial relationships among patches, this sequence is fed into another four layers of multi-head self-attention modules with positional encoding.

Afterward, the updated features are decoupled into three outputs. For the patch feature $F$, we use the saliency token $T_{s}^{g}$ and the contour token $T_{c}^{g}$ as the key and value ($K$, $V$), respectively, while taking the patch feature as the query $Q$. Through a Sigmoid-activated attention ($QK^{T} \to Sigmoid$) and a residual connection with the original patch feature $F$, we obtain the task-enhanced saliency feature $F_{s}$  and contour feature $F_{c}$. 
Meanwhile, the two updated global task tokens are passed through their respective LayerNorm and Feed-Forward Network (FFN) layers to generate the final global saliency token $T_{s}^{g}$ and global contour token $T_{c}^{g}$.


\subsection{Dual-branch Prior-guided Progressive Decoder (DPPD)}
Although GTC provides global semantics, the decoder still lacks explicit guidance from panoramic geometry and salient structures, limiting region discrimination and boundary precision. To address this, we propose DPPD, which injects priors into decoding to guide context aggregation and progressively refine predictions.



\subsubsection{Overlap-normalized Reverse T2T Upsampling (ON-rT2T)}
Reverse T2T upsamples tokens by folding projected $k\times k$ patches into a 2D map with stride $s$ ($s< k$). However, the $Fold$ operation directly sums overlapping regions. This uneven accumulation creates a spatial amplitude bias, degrading high-frequency boundary localization.

To address this, we propose ON-rT2T. Specifically, folding the projected tokens yields the accumulated map $U$. Simultaneously, applying the same $Fold$ to an all-ones tensor produces a coverage count map $D$. The final upsampled feature  $\hat{U}$ is obtained by element-wise division:
\begin{equation}
\small
U = \mathrm{Fold}(\mathrm{Proj}(T)),\quad
D = \mathrm{Fold}(\mathbf{1}),\quad
\hat{U} = \frac{U}{\max(D,1)} 
\end{equation}
where $max(D,1)$ avoids division by zero. ON-rT2T converts overlap summation into averaging, removing spatial bias without extra parameters. The upsampled features are then concatenated with the low-level encoder features and fused by linear projection to supplement fine local details.

\begin{figure}[t]
\centering
\includegraphics[scale=1]{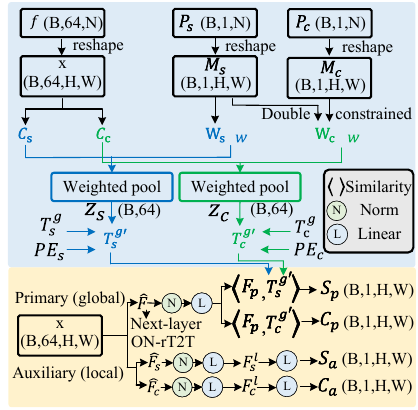}
\caption{\textbf {Overview of PGTR (top) and Dual Prediction (bottom). In PGTR, 
blue and green denote the saliency and contour branches, respectively. In Dual Prediction, the primary branch predicts via similarity with the updated tokens, while the auxiliary branch predicts directly from the task-enhanced features. 
}}
\label{fig5}
\end{figure}

\subsubsection{Prior-guided Token Refinement (PGTR)}
After each upsampling step, guided by three priors associated with panoramic geometry and salient-object structure, PGTR replaces the costly token-to-pixel self-attention with $O(N)$ weighted pooling. 
Specifically, as shown in Fig. \ref{fig5}, given the upsampled spatial feature $f$ and the saliency and contour coarse predictions from the previous stage $P_{s}$ and $P_{c}$, we reshape them into 2D maps $x$, $M_{s}$, and $M_{c}$, respectively. Here, $x\in\mathbb{R}^{B\times 64\times H\times W}$ denotes the spatial feature, and $\mathrm{M_{s}},\mathrm{M_{c}}\in\mathbb{R}^{B\times 1\times H\times W}$ denote the prior maps. PGTR consists of two branches, \textit{i.e.}, saliency (blue) and contour (green) branches, each guided by the following three types of priors.

\textbf{Foreground contrast prior.}
In panoramic images, salient objects tend to differ more from their half-circumference region than from adjacent surroundings, which
holds for 68.5\% and 72.1\% of the 360-SOD and 360-SSOD training images in CIELab histogram distance.
Based on this prior, the saliency branch rolls the feature by half the width and subtracts it from the original feature to obtain the saliency contrast feature $C_{s}$:
\begin{equation}
\mathrm{x_{\text{anti}}}=\mathrm{Roll}(x,\tfrac{W}{2}),\quad
\mathrm{C_{s}}=x-\mathrm{x_{\text{anti}}}
\end{equation}
where $\mathrm{C_{s}}$ is later fused as a weak residual modulated by a
learnable $\alpha_{sal}$ (initialized to $0$) and the gates
$g_{s},g_{c}$ (defined later in Eq.\ref{eq13}, Eq.\ref{eq15}).
For the contour branch, we extract the edges and apply the same half-width shift to yield the contour contrast feature $C_{c}$:
\begin{equation}
\mathrm{E}=\mathrm{Edge}(x),\quad
\mathrm{E_{\text{anti}}}=\mathrm{Roll}(\mathrm{E},\tfrac{W}{2}) 
\end{equation}
\begin{equation}
\mathrm{C_{c}}=\mathrm{E}+\alpha_{con}(\mathrm{E}-\mathrm{E_{\text{anti}}})
\end{equation}
where $\mathrm{Roll}(\cdot,\tfrac{W}{2})$ and $\mathrm{Edge}(\cdot)$ denote the half-width shift and edge extraction, respectively. Since contours are sparse, we formulate this contrast term as a weak residual controlled by $\alpha_{con}$ (initialized to $0$) to ensure stable early training.

\textbf{Sal-con prior.} This prior guides context aggregation toward structurally informative regions. For the saliency branch, the weights are directly obtained from the coarse saliency prediction, \textit{i.e.}, $W_s=\mathrm{M_{s}}$. For the contour branch, to exclude irrelevant high-frequency textures, we take the maximum of the coarse contour prediction and the saliency boundary:
\begin{equation}
W_c=\max\big(\mathcal{B}(\mathrm{M_{s}}),\mathrm{M_{c}}\big)
\end{equation}
where $\mathcal{B}(\cdot)$ denotes the panoramic boundary extractor, and $\max(\cdot)$ denotes the element-wise maximum operation.

\textbf{Pano-distortion prior.} Since ERP severely stretches polar regions, equal aggregation overweights distorted areas. Thus, we introduce a latitude-aware weight $w$:
\begin{equation}
w=\max\big(\cos(\tfrac{\pi y}{2}),0.25\big)
\end{equation}
where $y\in[-1,1]$ denotes the normalized latitude coordinate. $\cos(\tfrac{\pi y}{2})$ equals $1$ at the equator ($y=0$) and decays to $0$ toward the poles ($y=\pm 1$). To avoid excessively small weights in the polar regions, we truncate it to a minimum of $0.25$.

Guided by these priors, each branch uses $W_{s}/W_{c}\times w	$ as the final spatial weight for weighted average pooling. 
The saliency branch pools $x$ and its half-width-shifted feature $\mathrm{x_{\text{anti}}}$ separately and fuses them with a weak residual, while the contour branch directly pools $\mathrm{C_{c}}$:
\begin{equation}
\mathrm{Z_{\text{fg}}}=\frac{\sum_{h,w}(W_s w)\odot x}{\sum_{h,w}(W_s w)},\quad
\mathrm{Z_{\text{anti}}}=\frac{\sum_{h,w}(W_s w)\odot \mathrm{x_{\text{anti}}}}{\sum_{h,w}(W_s w)}
\end{equation}
\begin{equation}
\mathrm{Z_{s}}=\mathrm{Z_{\text{fg}}}+\alpha_{sal}(\mathrm{Z_{\text{fg}}}-\mathrm{Z_{\text{anti}})}
\end{equation}
\begin{equation}
\mathrm{Z_{c}}=\frac{\sum_{h,w}(W_c w)\odot \mathrm{C_{c}}}{\sum_{h,w}(W_c w)}
\end{equation}
where the summation is over all spatial positions $(h,w)$ and $\odot$ denotes element-wise multiplication. The denominator is used for weight normalization. $\alpha_{sal}$ is initialized to zero and introduces the contrast context as a learnable weak residual. The resulting global contexts $\mathrm{Z_{s}},\mathrm{Z_{c}}\in\mathbb{R}^{B\times 64}$ are task-relevant global representations at this scale.

Finally, we inject the global representations and positional encodings into the corresponding previous-stage global tokens, and update the tokens through residual connections:
\begin{equation}
\mathrm{T_{s}^{g\prime} }=\mathrm{T_{s}^{g}}+\phi_s\big(\mathrm{T_{s}^{g}}+\mathrm{PE_{s}}+\psi_s(\mathrm{Z_{s}})\big)
\end{equation}
\begin{equation}
\mathrm{T_{c}^{g\prime} }=\mathrm{T_{c}^{g}}+\phi_c\big(\mathrm{T_{c}^{g}}+\mathrm{PE_{c}}+\psi_c(\mathrm{Z_{c}})\big)
\end{equation}
where $\mathrm{T_{s}^{g}}$ and $\mathrm{T_{c}^{g}}$ denote the previous-stage global tokens, $\mathrm{PE_{s}}$ and $\mathrm{PE_{c}}$ are positional encodings, and $\phi_{{s,c}}$ and $\psi_{{s,c}}$ denote linear projections. The updated global tokens $\mathrm{T_{s}^{g\prime}}$, $\mathrm{T_{c}^{g\prime}}$ and spatial features are fed into the dual prediction.

\subsubsection{Dual prediction}
As shown in the lower part of Fig. \ref{fig5}, DPPD performs dual-branch prediction at each scale from the PGTR outputs. The primary branch uses the updated global tokens as task representations and fuses the original feature with gated contrast-enhanced features to obtain the enhanced primary feature:
\begin{equation}
\mathrm{\hat{F}}=x+0.5\cdot g_{s}\odot \mathrm{C_{s}}+0.5\cdot  g_{c}\odot \mathrm{C_{c}}
\label{eq13}
\end{equation}
where $\odot$ denotes element-wise multiplication, $g_{s}$/$g_{c}$ are saliency/contour gates controlling the enhancement strength. $\mathrm{\hat{F}}$ is normalized and projected to obtain the primary-branch feature $\mathrm{F_{p}}$, which computes position-wise similarities (denoted as $\left \langle  \right \rangle $ ) with the updated global tokens:
\begin{equation}
\mathrm{S_{p}}=  \left \langle \mathrm{F_{p}} , \mathrm{T_{s}^{g\prime}} \right \rangle ,\quad
\mathrm{S_{c}}= \left \langle \mathrm{F_{p}} , \mathrm{T_{c}^{g\prime}} \right \rangle    \end{equation}
here we obtain the primary-branch saliency prediction $\mathrm{S_{p}}$ and contour prediction $\mathrm{S_{c}}$. Meanwhile, $\mathrm{\hat{F}}$ is passed to the next ON-rT2T stage for progressive high-resolution refinement.

The auxiliary branch performs dense prediction directly from task-specific enhanced features, without using global tokens. Specifically, the original feature $x$ is added to each gated contrast enhancement to obtain the saliency-enhanced feature $\mathrm{\hat{F}_{s}}$ and the contour-enhanced feature $\mathrm{\hat{F}_{c}}$:
\begin{equation}
\mathrm{\hat{F}_{s}}=x+g_s\odot \mathrm{C_{s}},\quad
\mathrm{\hat{F}_{c}}=x+g_c\odot \mathrm{C_{c}}
\label{eq15}
\end{equation}
they are normalized and projected into local features $\mathrm{F_{s}^{l}}$ and $\mathrm{F_{c}^{l}}$, then classified linearly to yield the auxiliary predictions $\mathrm{S_{a}}$ and $\mathrm{C_{a}}$, complementing the primary branch with local details.

Note that the 1/16 scale differs slightly. Lacking a previous coarse prediction, it directly uses the GTC outputs (Fig. \ref{fig4}). Specifically, the primary branch matches the patch feature $F$ with global tokens $\mathrm{T_{s}^{g}}$ and $\mathrm{T_{c}^{g}}$, while the auxiliary branch predicts from the task-enhanced features $\mathrm{F_{s}}$ and $\mathrm{F_{c}}$. Starting from the 1/8 scale, each stage follows the standard pipeline—ON-rT2T, PGTR, and dual prediction—guided by the previous coarse prediction.

\begin{figure}[t]
\centering
\includegraphics[scale=1]{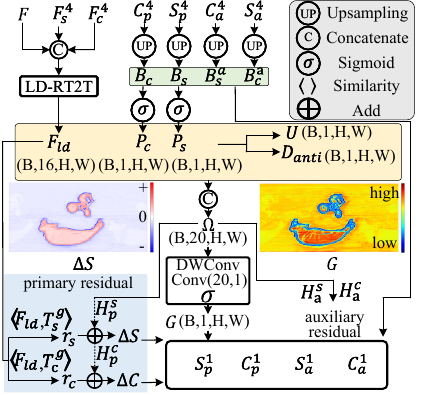}
\caption{\textbf {Illustration of PCRD. Taking the coarse predictions as the base, PCRD modulates the residuals with the predicted gate to achieve boundary refinement, yielding high-quality full-resolution predictions.}}
\label{fig6}
\end{figure}

\subsubsection{Probability-context Residual Decoder (PCRD)}
After three refinement stages, direct bilinear upsampling from 1/4-scale blurs salient boundaries. To efficiently recover details, we propose PCRD (Fig. \ref{fig6}). 
It takes inputs from the 1/4 stage, 
spatial features ($\mathrm{F}$, $\mathrm{F_{s}^{4}}$, $\mathrm{F_{c}^{4}}$, collectively $f_{a}^{4} $ in Fig. \ref{fig2}), 
primary predictions ($\mathrm{S_{p}^{4}}$, $\mathrm{C_{p}^{4}}$), 
auxiliary predictions ($\mathrm{S_{a}^{4} }$, $\mathrm{C_{a}^{4}}$),
and updated global tokens ($\mathrm{T_{s}^{g}}$, $\mathrm{T_{c}^{g}}$ , collectively $f_{p}^{4} $ in Fig. \ref{fig2}).

We first concatenate the three 1/4-scale spatial features and  upsample them via a $16$-channel low-dimensional reverse T2T (LD-RT2T), producing the full-scale feature $\mathrm{F_{\text{ld}}}\in\mathbb{R}^{B\times 16\times H\times W}$. This preserves structural details for boundary refinement with low computational cost. Meanwhile, the coarse predictions are bilinearly upsampled into full-scale logit bases ($\mathrm{B_{s}}$, $\mathrm{B_{c}}$ for primary branch and $\mathrm{B_{s}^{a}}$, $\mathrm{B_{c}^{a}}$ for auxiliary branch ) to serve as references for residual refinement.

To stabilize gate and residual learning, we build the context in the probability domain. The primary logit bases are mapped by Sigmoid to $\mathrm{P_{s}}$ and $\mathrm{P_{c}}\in[0,1]$, and two priors, uncertainty $\mathrm{U}$ and saliency difference $\mathrm{D_{anti}}$, are derived from $\mathrm{P_{s}}$:
\begin{equation}
\mathrm{U}=1-\big|2\cdot \mathrm{P_{s}}-1\big|,\quad
\mathrm{D_{anti}}=\mathrm{P_{s}}-\mathrm{Roll}(\mathrm{P_{s}},\tfrac{W}{2})
\end{equation}
where $\mathrm{U}$ is high near uncertain regions (near contours, $\mathrm{P_{s}} \approx 0.5$) and low near confident regions (background or object interiors,  $\mathrm{P_{s}} \approx 0,1$), thus indicating where refinement is needed. $\mathrm{D_{anti}}$  captures the saliency contrast between the foreground and its half-width-shifted region. We concatenate these probability-domain maps with the low-dimensional feature to form the context $\mathrm{\Omega}$:
\begin{equation}
\Omega=\mathrm{Concat_{c}}(\mathrm{F_{\text{ld}}}, \mathrm{P_{s}},\mathrm{P_{c}},\mathrm{U},\mathrm{D_{anti}})\in\mathbb{R}^{B\times 20\times H\times W}
\end{equation}
where $\Omega$ has $20$ channels ($16+4$). Its bounded values ensure numerically stable inputs for both gate and residual prediction. In the primary branch, the residuals combine context residuals from heads $\mathrm{H_{p}^{s}}(\mathrm{\Omega})$ and $\mathrm{H_{p}^{c}}(\mathrm{\Omega})$ with token responses $\mathrm{r_{s}}$ and $\mathrm{r_{c}}$, computed by similarity ($\left \langle  \right \rangle$) between $\mathrm{F_{\text{ld}}}$ and the previous global tokens $\mathrm{T_{s}^{g}}$ and $\mathrm{T_{c}^{g}}$. The summation yields the final residuals $\mathrm{\Delta S}$ and $\mathrm{\Delta C}$:
\begin{equation}
\mathrm{\Delta S}=\mathrm{H_{p}^{s}}(\mathrm{\Omega})+\mathrm{r_{s}},\quad
\mathrm{\Delta C}=\mathrm{H_{p}^{c}}(\mathrm{\Omega})+\mathrm{r_{c}}
\end{equation}
The auxiliary residuals are directly predicted from the context without token responses, using heads $\mathrm{H_{a}^{s}}(\mathrm{\Omega})$ and  $\mathrm{H_{a}^{c}}(\mathrm{\Omega})$. 

The gate $\mathrm{G}$ is generated from the context via a panoramic depthwise separable convolution ($\mathrm{DWConv}$), a $1\times1$ convolution ($\mathrm{Conv}(20,1)$), and a Sigmoid activation:
\begin{equation}
\mathrm{G}=\sigma\big(\mathrm{Conv}_{1\times1}(\mathrm{DWConv}(\mathrm{\Omega}))\big)\in[0,1]^{B\times 1\times H\times W}
\end{equation}
The modulation of the residuals by the gate exhibits explicit boundary sensitivity. 
As visualized by $\mathrm{\Delta S}$ and $\mathrm{G}$ in Fig. \ref{fig6}, $\mathrm{\Delta S}$ exhibits small positive values inside objects, large positive/negative values along inner/outer boundaries, and near-zero values in the background. Meanwhile, $\mathrm{G}$ is strictly positive, with stronger responses inside objects and weaker responses near boundaries.
After multiplying  $\mathrm{G}$ and $\mathrm{\Delta S}$, the gate amplifies positive residuals inside objects to enhance saliency, while its weaker values at boundaries moderately constrain the positive/negative residuals, jointly sharpening contours and suppressing boundary leakage.
Finally, the gate-modulated residuals are added to the base logits to obtain full-resolution primary- and auxiliary-branch predictions:
\begin{equation}
\mathrm{S_{p}^{1}}=\mathrm{B_{s}}+\mathrm{G}\cdot \mathrm{\Delta S} ,\quad
\mathrm{C_{p}^{1}}=\mathrm{B_{c}}+\mathrm{\Delta C} 
\end{equation}
\begin{equation}
\mathrm{S_{a}^{1}}=\mathrm{B_{s}^{a}}+\mathrm{G}\cdot \mathrm{H_{a}^{s}}(\mathrm{\Omega}) ,\quad
\mathrm{C_{a}^{1}}=\mathrm{B_{c}^{a}}+\mathrm{H_{a}^{c}}(\mathrm{\Omega})
\end{equation}
Here, $\mathrm{G}$ is applied only to the saliency branch, as contours already represent boundaries and require no additional filtering.

\subsection{Loss Function}
\textbf{BCE Loss.} Binary cross-entropy (BCE) measures pixel-level discrepancy and serves as the fundamental supervision signal. We apply BCE to both saliency and contour predictions across all $S$ scales, using downsampled ground truth (GT) for supervision. The multi-scale BCE loss $\mathcal{L}_{bce}$ is defined as:
\begin{equation}
\mathcal{L}_{bce}=\sum_{i=1}^{S} w_i\Big[{BCE}({P}_{sal}^{i},{G}_{sal}^{i})+{BCE}({P}_{con}^{i},{G}_{con}^{i})\Big]
\end{equation}
where ${P}_{sal}^{i}$ and ${P}_{con}^{i}$ denote the saliency and contour predictions at the $i$-th scale, respectively, while ${G}_{sal}^{i}$ and ${G}_{con}^{i}$ are the corresponding downsampled GT. $w_i$ is the scale weight, empirically set to $w_1{=}1.0$, $w_2{=}0.8$, $w_3{=}0.5$, and $w_4{=}0.5$ from high to low resolutions.

\textbf{IoU Loss.}
Since pixel-wise BCE may overlook small objects under foreground-background imbalance, we further apply IoU loss to saliency and contour predictions across all $S$ scales to enforce regional overlap consistency. The multi-scale IoU loss $\mathcal{L}_{iou}$ is defined as:
\begin{equation}
\mathcal{L}_{iou}=\sum_{i=1}^{S} w_i\Big[{IoU}({P}_{sal}^{i},{G}_{sal}^{i})+{IoU}({P}_{con}^{i},{G}_{con}^{i})\Big]
\end{equation}
where the scale weights $w_i$ are the same as in $\mathcal{L}_{bce}$. The base loss $\mathcal{L}_{base}$ combines BCE and IoU, and the auxiliary counterpart $\mathcal{L}_{base}^{aux}$ is defined similarly.
\begin{equation}
\mathcal{L}_{base}=\mathcal{L}_{bce}+\mathcal{L}_{iou}
\end{equation}


\textbf{Align Loss.} 
Neither BCE nor IoU explicitly constrains the intensity distribution consistency between the prediction and GT. To address this, we introduce the alignment loss $\mathcal{L}_{align}$ to align the pixel-wise deviation directions from the global average, thereby improving structural consistency. Given the full-resolution saliency prediction ${P}_{sal}^{1}$ and GT ${G}_{sal}^{1}$, we first mean-center them to obtain their deviations:
\begin{equation}
\varphi_P={P}_{sal}^{1}-\mu_P,\qquad \varphi_G={G}_{sal}^{1}-\mu_G
\end{equation}
where $\mu_P$ and $\mu_G$ are the global means of ${P}_{sal}^{1}$ and ${G}_{sal}^{1}$, respectively. We then compute the pixel-wise alignment between their deviations and normalize it.
\begin{equation}
\xi=\dfrac{2\cdot\varphi_P\cdot\varphi_G}{\varphi_P^2+\varphi_G^2+\epsilon},\quad
\phi=\frac{(\xi+1)^2}{4}
\end{equation}
Here $\epsilon$ is a small constant to avoid division by zero. $\xi\in[-1,1]$ is utilized to measure the directional consistency between the prediction and GT deviations: it approaches $1$ when they are in the same direction and approaches $-1$ when they are opposite. $\phi$ maps to $[0,1]$ to amplify well-aligned regions. The alignment loss $\mathcal{L}_{align}$ is defined as:
\begin{equation}
\mathcal{L}_{align}=1-\frac{\sum_{hw}^{}  \phi} {H\times W} 
\end{equation}
$\mathcal{L}_{align}$ captures global intensity distribution consistency, compensating for the pixel-wise and foreground-only limitations of BCE and IoU. We apply it to the full-resolution primary and auxiliary saliency predictions. The total loss $\mathcal{L}_{total}$ is:
\begin{equation}
\mathcal{L}_{total}=\mathcal{L}_{base}+\mathcal{L}_{base}^{aux}+\mathcal{L}_{align}+\mathcal{L}_{align}^{aux}
\end{equation}

\begin{table*}[t]
\centering
\begin{threeparttable}
\footnotesize   
\caption{Quantitative comparison of SphereSOD with other methods on the 360-SOD dataset in terms of $S_m$, MAE, F-Measure, and E-Measure. $\uparrow$ ($\downarrow$) indicates higher (lower) is better. The top three results are highlighted in \textcolor{red}{red}, \textcolor{green}{green}, and \textcolor{blue}{blue}.}
\label{tab_360sod}
\setlength{\tabcolsep}{2pt}
\renewcommand{\arraystretch}{0.9}
\begin{tabular}{c|c|c|cccccccc|ccccccccccc}
\toprule
\multirow{3}{*}{\textbf{Dataset}} & \multirow{3}{*}{\textbf{Metric}} & \multirow{3}{*}{\textbf{Ours}} 
& \multicolumn{8}{c|}{\textbf{360$^\circ$ Models}} & \multicolumn{11}{c}{\textbf{2D Models}} \\
\cmidrule(lr){4-11} \cmidrule(lr){12-22}
& & & SIHE & SCFA & MPFR & LD & DDS & FANet & DAT & HUA & BIPG & PG & GFI & SCW & BPFI & GCPA & HVP & MI & ACCo & LDF & SCRN \\
& & & {\cite{r41}} & {\cite{r42}} & {\cite{r23}} & {\cite{r43}} & {\cite{r20}} & {\cite{r22}} & {\cite{r25}} & {\cite{r88}} & {\cite{r44}} & {\cite{r45}} & {\cite{r46}} & {\cite{r47}} & {\cite{r48}} & {\cite{r49}} & {\cite{r50}} & {\cite{r51}} & {\cite{r52}} & {\cite{r53}} & {\cite{r54}} \\
\cmidrule(lr){1-22}
\multirow{8}{*}{\rotatebox{90}{\textbf{360-SOD Dataset}}} 
& $S_m \uparrow$ & \textcolor{red}{.889}   & \textcolor{green}{.871} & \textcolor{green}{.871} & .842 & .768 & .799 & .826 & .849 & .847 & .811 & .750 & .831 & .830 & .788 & .674 & .709 & .719 & .770 & .815 & .792 \\
& $MAE \downarrow$ & \textcolor{red}{.014} & \textcolor{green}{.017} & .018 & .019 & .029 & .023 & .021 & \textcolor{green}{.017} & \textcolor{green}{.017}   & .024  & .030 & .021 & .021 & .024 & .040 & .052 & .050 & .025 & .023 & .024 \\
\cmidrule(lr){2-22}
& $adpE \uparrow$ & \textcolor{red}{.936}  & \textcolor{green}{.932} & \textcolor{blue}{.923} & .890 & .858 & .854 & .883 & - & - & .884  & .755 & .895 & .886 & .829 & .791 & .679 & .696 & .866 & .827 & .834 \\
& $meanE \uparrow$ & \textcolor{red}{.935}   & \textcolor{green}{.931} & \textcolor{blue}{.925} & .875 & .844 & .865 & .873 & .907 & .908 & .885  & .745 & .894 & .886 & .819 & .770 & .734 & .770 & .768 & .851 & .846 \\
& $maxE \uparrow$ & \textcolor{red}{.939} & \textcolor{green}{.937} & \textcolor{blue}{.930} & .885 & .866 & .904 & .900 & .919 & - & .890  & .786 & .900 & .887 & .865 & .785 & .842 & .824 & .869 & .877 & .877 \\
\cmidrule(lr){2-22}
& $adpF \uparrow$ & \textcolor{red}{.831} & \textcolor{green}{.796} & \textcolor{blue}{.793} & .745 & .617 & .638 & .700 & - & - & .717  & .648 & .735 & .752 & .603 & .491 & .423 & .443 & .717 & .633 & .615 \\
& $meanF \uparrow$ & \textcolor{red}{.836}  & \textcolor{green}{.813} & \textcolor{blue}{.808} & .755 & .641 & .695 & .748 & .774 & .786 & .727  & .629 & .753 & .754 & .672 & .501 & .545 & .545 & .676 & .715 & .675 \\
& $maxF \uparrow$ & \textcolor{red}{.848} & \textcolor{green}{.830} & \textcolor{blue}{.824} & .765 & .656 & .722 & .770 & .793 & - & .740  & .646 & .769 & .759 & .708 & .508 & .655 & .655 & .735 & .740 & .707 \\
\bottomrule
\end{tabular}
\end{threeparttable}
\end{table*}

\begin{table*}[t]
\centering
\begin{threeparttable}
\footnotesize 
\caption{Quantitative comparison on the 360-SSOD dataset. Notations follow Table~\ref{tab_360sod}.}
\label{tab_360ssod}
\setlength{\tabcolsep}{3pt}
\renewcommand{\arraystretch}{0.9}
\begin{tabular}{c|c|c|ccccc|ccccccccccc}
\toprule
\multirow{3}{*}{\textbf{Dataset}} & \multirow{3}{*}{\textbf{Metric}} & \multirow{3}{*}{\textbf{Ours}} 
& \multicolumn{5}{c|}{\textbf{360$^\circ$ Models}} & \multicolumn{11}{c}{\textbf{2D Models}} \\
\cmidrule(lr){4-8} \cmidrule(lr){9-19}
& & & SIHE & SCFA  & LD  & FANet & DAT & BIPG & PG & GFI & SCW & BPFI & GCPA & HVP & MI & ACCo & LDF & SCRN \\
& & & {\cite{r41}} & {\cite{r42}}  & {\cite{r43}} & {\cite{r22}} & {\cite{r25}} & {\cite{r44}} & {\cite{r45}} & {\cite{r46}} & {\cite{r47}} & {\cite{r48}} & {\cite{r49}} & {\cite{r50}} & {\cite{r51}} & {\cite{r52}} & {\cite{r53}} & {\cite{r54}} \\
\cmidrule(lr){1-19}
\multirow{8}{*}{\rotatebox{90}{\textbf{360-SSOD Dataset}}} 
& $S_m \uparrow$ & \textcolor{red}{.795} & \textcolor{blue}{.788} & \textcolor{green}{.791} & .756 & .717 & .770 & .760 & .712 & .767 & .760 & .766 & .741 & .773 & .732 & .747 & .752 & .748 \\
& $MAE \downarrow$ & \textcolor{red}{.026} & \textcolor{blue}{.028} & .029 & .034 & .039 & \textcolor{red}{.026} & .030 & .041 & .034 & .029 & .030 & .035 & .030 & .059 & .031 & .033 & .031 \\
\cmidrule(lr){2-19}
& $adpE \uparrow$ & \textcolor{red}{.879} & \textcolor{green}{.876} & \textcolor{blue}{.867} & .841 & .717 & - & .864 & .793 & .842 & .856 & .773 & .836 & .782 & .660 & .797 & .713 & .766 \\
& $meanE \uparrow$ & \textcolor{red}{.882} & \textcolor{green}{.871} & \textcolor{blue}{.860} & .845 & .727 & .832 & .817 & .731 & .829 & .813 & .795 & .811 & .800 & .778 & .758 & .749 & .780 \\
& $maxE \uparrow$ & \textcolor{red}{.889} & \textcolor{green}{.886} & \textcolor{blue}{.870} & .863 & .735 & .865 & .855 & .790 & .848 & .854 & .851 & .837 & .854 & .819 & .853 & .854 & .851 \\
\cmidrule(lr){2-19}
& $adpF \uparrow$ & \textcolor{red}{.657} & \textcolor{green}{.558} & \textcolor{blue}{.551} & .461 & .516 & - & .523 & .457 & .516 & .523 & .420 & .468 & .435 & .344 & .432 & .391 & .414 \\
& $meanF \uparrow$ & \textcolor{red}{.682} & \textcolor{blue}{.567} & .560 & .492 & .520 & \textcolor{green}{.644} & .517 & .425 & .526 & .511 & .483 & .469 & .500 & .453 & .465 & .454 & .470 \\
& $maxF \uparrow$ & \textcolor{red}{.701} & \textcolor{blue}{.576} & .570 & .511 & .532 & \textcolor{green}{.657} & .524 & .452 & .536 & .518 & .512 & .479 & .529 & .513 & .503 & .511 & .500 \\
\bottomrule
\end{tabular}
\end{threeparttable}
\end{table*}

\begin{table*}[t]
\centering
\begin{threeparttable}
\footnotesize 
\caption{Quantitative comparison on the ODI-SOD dataset. Notations follow Table~\ref{tab_360sod}.}
\label{tab_odi-sod}
\setlength{\tabcolsep}{3pt}
\renewcommand{\arraystretch}{0.9}
\begin{tabular}{c|c|c|cccccc|ccccccccccc}
\toprule
\multirow{3}{*}{\textbf{Dataset}} & \multirow{3}{*}{\textbf{Metric}} & \multirow{3}{*}{\textbf{Ours}} 
& \multicolumn{6}{c|}{\textbf{360$^\circ$ Models}} & \multicolumn{11}{c}{\textbf{2D Models}} \\
\cmidrule(lr){4-9} \cmidrule(lr){10-20}
& & & SIHE & View  & SCFA  & DDS & FANet & DAT   & BIPG & PG & GFI & SCW & BPFI & GCPA & HVP & MI & ACCo & LDF & SCRN \\
& & & {\cite{r41}} & {\cite{r24}}  & {\cite{r42}} & {\cite{r20}} & {\cite{r22}} & {\cite{r25}} & {\cite{r44}} & {\cite{r45}} & {\cite{r46}} & {\cite{r47}} & {\cite{r48}} & {\cite{r49}} & {\cite{r50}} & {\cite{r51}} & {\cite{r52}} & {\cite{r53}} & {\cite{r54}} \\
\cmidrule(lr){1-20}
\multirow{8}{*}{\rotatebox{90}{\textbf{ODI-SOD Dataset}}} 
& $S_m \uparrow$ & \textcolor{red}{.887} & \textcolor{green}{.846} & .831 & \textcolor{blue}{.840} & .791 & .730 & .828  & .815 & .808 & .779 & .814 & .822 & .826 & .732 & .803 & .681 & .802 & .817 \\
& $MAE \downarrow$ & \textcolor{red}{.026} & \textcolor{blue}{.036} & \textcolor{green}{.035} & .037 & .045 & .050 & \textcolor{blue}{.036}  & .042 & .044 & .051 & .043 & .040 & .042 & .061 & .040 & .094 & .046 & .045 \\
\cmidrule(lr){2-20}
& $adpE \uparrow$ & \textcolor{red}{.922} & \textcolor{green}{.887} & \textcolor{blue}{.886} & .878 & .808 & .778 & .866 & .864 & .854 & .816 & .862 & .870 & .845 & .795 & .853 & .616 & .839 & .816 \\
& $meanE \uparrow$ & \textcolor{red}{.918} & \textcolor{green}{.888} & - & \textcolor{blue}{.880} & - & .771 & .858  & .861 & .851 & .801 & .852 & .861 & .850 & .787 & .869 & .725 & .821 & .839 \\
& $maxE \uparrow$ & \textcolor{red}{.924} & \textcolor{green}{.894} & - & \textcolor{blue}{.886} & - & .790 & .884  & .867 & .857 & .810 & .859 & .869 & .862 & .795 & .881 & .865 & .862 & .867 \\
\cmidrule(lr){2-20}
& $adpF \uparrow$ & \textcolor{red}{.846} & \textcolor{green}{.777} & .759 & \textcolor{blue}{.763} & .630 & .608 & .711  & .738 & .715 & .696 & .744 & .759 & .702 & .607 & .708 & .439 & .690 & .664 \\
& $meanF \uparrow$ & \textcolor{red}{.850} & \textcolor{green}{.789} & - & \textcolor{blue}{.776} & - & .610 & .754  & .744 & .727 & .692 & .745 & .760 & .738 & .616 & .757 & .551 & .721 & .726 \\
& $maxF \uparrow$ & \textcolor{red}{.859} & \textcolor{blue}{.808} & \textcolor{green}{.822} & .793 & .761 & .632 & .778  & .759 & .743 & .700 & .753 & .770 & .757 & .627 & .779 & .754 & .748 & .754 \\
\bottomrule
\end{tabular}
\end{threeparttable}
\end{table*}

\section{Experiments}
\subsection{Datasets and Implementation Details}
We evaluate our method on three 360$^{\circ }$ SOD benchmarks: 360-SOD \cite{r20} (400 train / 100 test), 360-SSOD \cite{r21} (850 train / 255 test), and ODI-SOD \cite{r24} (4263 train / 2000 test).

All experiments are conducted on a single GTX 3090 GPU with a batch size of 2. We use the Adam optimizer \cite{r37} with an initial learning rate of 0.0001, decayed by a factor of 0.1. Following \cite{r25}, contour maps are generated from GT saliency maps for contour supervision. We use Swin-S \cite{r38} as the backbone and implement the model in PyTorch \cite{r40}.


For data augmentation, each ERP image is resized to $1024\times512$. We apply random horizontal flipping (prob. 0.5) and circular rolling (prob. 0.7), where the roll offset is uniformly sampled from $[5\%, 95\%]$ of the width. Color jittering (prob. 0.5) further changes brightness, contrast, and saturation by $\pm0.3$, and hue by $\pm0.05$.

\subsection{Evaluation Metrics}
We assess our model with four standard saliency detection metrics: S-measure ($S_{m}$) \cite{r33}, MAE \cite{r34}, F-measure ($F_{\beta }$) \cite{r35}, and E-measure ($E_{m}$) \cite{r36}. Higher $S_{m}$/$F_{\beta }$/$E_{m}$ and lower MAE are better. $S_{m}$ evaluates structural similarity:
\begin{equation}
S_{m} =\alpha S_{o} +(1-\alpha )S_{r} 
\end{equation}
where $S_{o}$ and $S_{r}$ denote object-aware and region-aware structural similarity, and $\alpha$ is usually
set to 0.5 by default. MAE measures the average pixel-wise discrepancy between the predicted saliency map $S$ and the ground truth $GT$:
\begin{equation}
MAE=\frac{1}{H\times W} \sum_{y=1}^{H} \sum_{x=1}^{W} \left | S(x,y)-GT(x,y) \right | 
\label{equ13}
\end{equation}
where $W$ and $H$ are the width and height of the image. $F_{\beta }$ summarizes the trade-off between precision $P$ and recall $R$:
\begin{equation}
F_{\beta } =\frac{(1+\beta ^{2} )\cdot P\cdot R}{\beta ^{2}\cdot P+R} 
\label{equ14}
\end{equation}
where $\beta ^{2}$ is set to 0.3 to emphasize precision. Following common practice, we report the adaptive/mean/max $F_{\beta }$. $E_{m}$ prediction–ground-truth agreement via an enhanced alignment matrix that fuses per-pixel matching with image-level statistics, capturing both local and global similarity:
\begin{equation}
E_{m}=\frac{1}{H\times W} \sum_{y=1}^{H} \sum_{x=1}^{W} \cdot \text{FM}(x,y)
\label{equ14}
\end{equation}
where FM is the enhanced alignment matrix computed from the saliency map and ground truth. We also report the adaptive/mean/max $E_{m}$. These metrics reflect complementary aspects of performance: structural similarity ($S_{m}$), absolute error (MAE), precision–recall trade-off ($F_{\beta }$), and alignment at both global and local levels ($E_{m}$).

\begin{figure*}[t]
\centering
\includegraphics[scale=0.30]{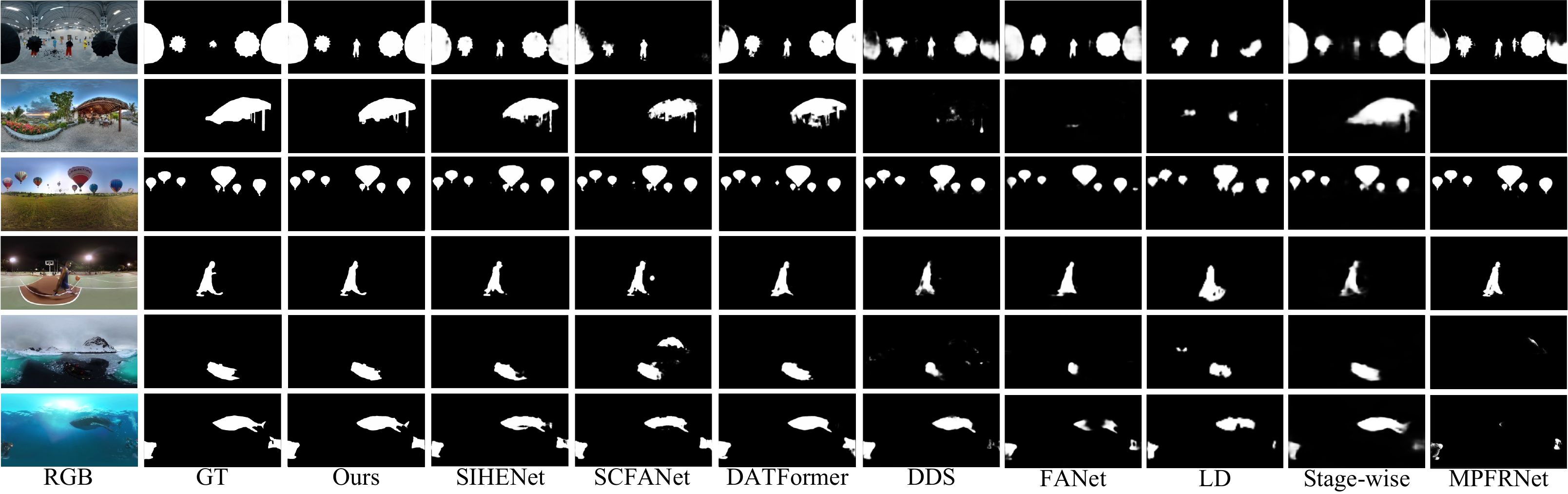}
\caption{\textbf { Qualitative comparison results of our method and other SOTA models on the 360-SOD dataset.}}
\label{fig7}
\end{figure*}

\begin{figure*}[t]
\centering
\includegraphics[scale=0.30]{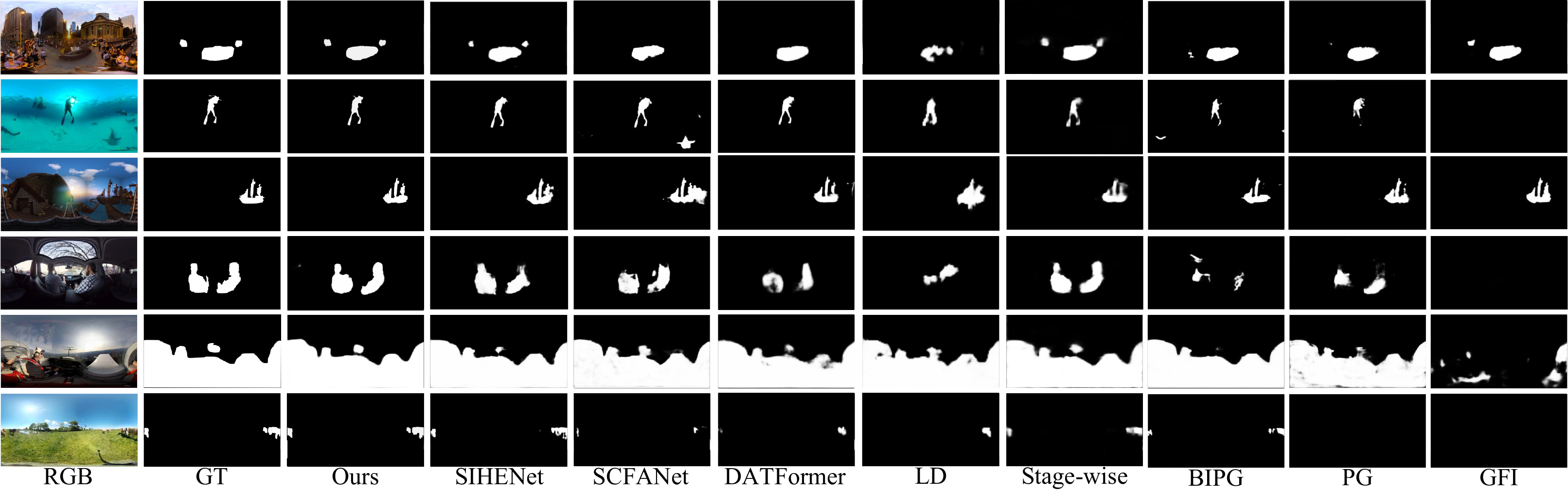}
\caption{\textbf { Qualitative comparison results of our method and other SOTA models on the 360-SSOD dataset.}}
\label{fig8}
\end{figure*}

\begin{figure*}[t]
\centering
\includegraphics[scale=0.33]{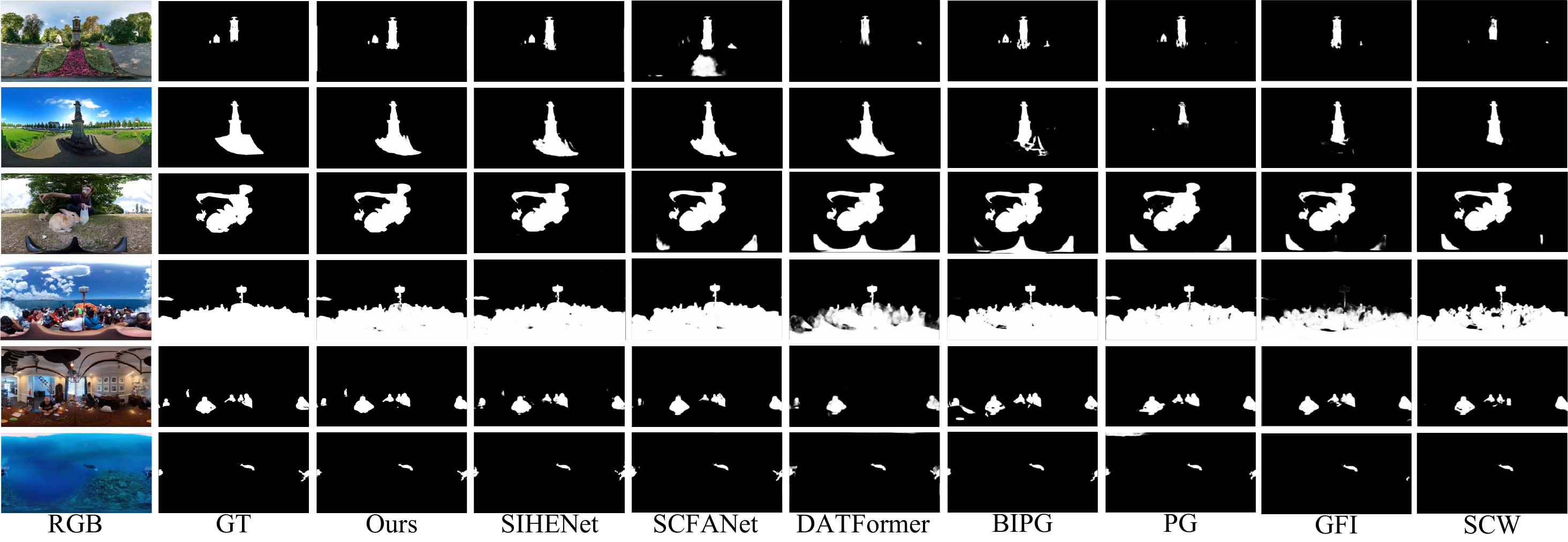}
\caption{\textbf { Qualitative comparison results of our method and other SOTA models on the ODI-SOD dataset.}}
\label{fig9}
\end{figure*}

\subsection{Comparison Results}
To validate our method, we compare it with SOTA methods on three datasets, including 360$^\circ$ SOD methods \cite{r41,r42,r23,r43,r20,r22,r25,r88,r24,r21} and 2D SOD methods \cite{r44,r45,r46,r47,r48,r49,r50,r51,r52,r53,r54}.
\subsubsection{Quantitative Analysis}
As shown in Tables \ref{tab_360sod}, \ref{tab_360ssod} and \ref{tab_odi-sod}, our method achieves the best performance across all metrics on the three benchmarks. Compared with the second-best methods, the gains in $S_m$, $MAE$, $adpE$, and $adpF$ are:
360-SOD: 2.07\%, 17.65\%, 4.29\%, and 4.40\%;
360-SSOD: 5.06\%, 7.14\%, 3.42\%, and 17.74\%;
ODI-SOD: 4.85\%, 25.71\%, 3.95\%, and 8.89\%, respectively.

\subsubsection{Qualitative Analysis}
\textbf{360-SOD dataset:} Fig. \ref{fig7} compares our saliency maps with other SOTA methods on 360-SOD. 
In multi-object scenes (Rows 1,3,6), our method more accurately identifies, locates, and segments all salient objects.
Under severe distortion (Row 2), it still robustly localizes the target. 
For detail-rich scenes (Rows 4,5), it better preserves object details and produces more accurate boundaries.

\textbf{360-SSOD dataset:} Fig. \ref{fig8} shows visual comparisons on 360-SSOD. In multi-object scenes (Rows 1,4,6), our method reliably detects and segments all salient objects. For complex-detail scenes (Rows 2,3), it achieves finer segmentation with clearer boundaries. In the large-object scene (Row 5), it also preserves object details while accurately delineating contours.

\textbf{ODI-SOD dataset:} Fig. \ref{fig9} shows visual comparisons on ODI-SOD. In scenes with both large and small objects (Rows 1,5,6), our method detects all targets and preserves fine local details. In detail-complex scenes (Rows 2,3), it effectively suppresses background interference and recovers accurate boundaries. In the large complex scene (Row 4), competing methods often produce internal holes or missed detections, whereas ours yields complete regions and precise contours.
\begin{table}[t]
\centering
\footnotesize
\renewcommand{\arraystretch}{0.9}
\caption{Complexity comparison against multi-projection fusion methods on the 360-SOD dataset.}
\label{tab_complexity}
\setlength{\tabcolsep}{3pt}
\begin{tabular}{lccccc}
\toprule
Method & FANet~\cite{r22} & MPFRNet~\cite{r23} & SCFANet~\cite{r42} & HUA~\cite{r88} & Ours \\
\midrule
FLOPs (G)  & 241.39 & 84.91  & 713.69 & 276.32 & 135.27 \\
Params (M) & 25.40  & 117.72 & 226.13 & 115.94 & 71.18  \\
\bottomrule
\end{tabular}
\end{table}

\subsubsection{Complexity Comparison}
To compare computational complexity, we report the FLOPs and parameters of our method and representative multi-projection fusion methods (results from~\cite{r88}) on 360-SOD, at $1024\times512$ with a batch size 1 (Table~\ref{tab_complexity}). 
Our method attains only 135.27G FLOPs (56.0\% of FANet~\cite{r22}, 19.0\% of SCFANet~\cite{r42} and 49.0\% of HUA~\cite{r88}), 71.18M parameters (far fewer than MPFRNet~\cite{r23}, SCFANet~\cite{r42} and HUA~\cite{r88} ). This is because our method discards extra projection transformations and couples panoramic geometry with salient structures for encoding and decoding on a single ERP, achieving markedly lower overhead while comprehensively outperforming these methods.

\begin{table}[t]
\centering
\footnotesize
\renewcommand{\arraystretch}{0.9}
\caption{Generalization to a weaker backbone (ResNet-50).}
\label{tab_backbone}
\setlength{\tabcolsep}{4pt}
\begin{tabular}{llcccc}
\toprule
Dataset & Method & $S_m\uparrow$ & MAE$\downarrow$ & mean$E\uparrow$ & mean$F\uparrow$ \\
 \midrule
\multirow{2}{*}{360-SOD}& Baseline     & 0.8099 & 0.0245 & 0.8666 & 0.7255 \\
 & Ours         & \textbf{0.8327} & \textbf{0.0208} & \textbf{0.8992} & \textbf{0.7640} \\
\addlinespace[2pt]
\multirow{2}{*}{360-SSOD}
 & Baseline    & 0.7383 & 0.0337 & 0.8223 & 0.5904 \\
 & Ours         & \textbf{0.7660} & \textbf{0.0289} & \textbf{0.8541} & \textbf{0.6220} \\
\addlinespace[2pt]
\multirow{2}{*}{ODI-SOD}
 & Baseline      & 0.7902 & 0.0541 & 0.8344 & 0.6979 \\
 & Ours         & \textbf{0.8305} & \textbf{0.0399} & \textbf{0.8775} & \textbf{0.7677} \\
\bottomrule
\end{tabular}
\end{table}

\subsubsection{Generalization across Backbones} 
To verify that the gain stems from our design rather than the Swin-S backbone, we replace it with a weaker ResNet-50 and retrain two variants under identical settings: a plain baseline (without PG-DSA and DPPD) and our full model. As shown in Table~\ref{tab_backbone}, the full model consistently surpasses the baseline on all three datasets, confirming that the improvement stems from the proposed design rather than the backbone.

\begin{table}[t]
\centering
\caption{Ablation study on 360-SOD and ODI-SOD datasets. Each component group is evaluated by removing or replacing the corresponding module. The best result per column is in \textbf{bold}.}
\label{tab_ablation}
\setlength{\tabcolsep}{2.5pt}
\renewcommand{\arraystretch}{0.95}
\resizebox{\columnwidth}{!}{%
\begin{tabular}{l cccc cccc}
\toprule
\multirow{2}{*}{Method} & \multicolumn{4}{c}{\textbf{360-SOD}} & \multicolumn{4}{c}{\textbf{ODI-SOD}} \\
\cmidrule(lr){2-5} \cmidrule(lr){6-9}
 & $S_m\uparrow$ & MAE$\downarrow$ & mean$E\uparrow$ & mean$F\uparrow$ & $S_m\uparrow$ & MAE$\downarrow$ & mean$E\uparrow$ & mean$F\uparrow$ \\
\midrule
\textbf{Ours} & \textbf{.889} & \textbf{.014} & \textbf{.935} & \textbf{.836} & \textbf{.887} & .026 & .918 & \textbf{.850} \\
\midrule
\multicolumn{9}{c}{\textit{Group 1: Modules}} \\
\midrule
w/o PG-DSA          & .875 & .015 & .923 & .825 & .871 & .031 & .909 & .823 \\
w/o DPPD            & .874 & .016 & .921 & .806 & .868 & .031 & .903 & .823 \\
Baseline (w/o both) & .859 & .017 & .916 & .804 & .849 & .040 & .883 & .783 \\
\midrule
\multicolumn{9}{c}{\textit{Group 2: Components}} \\
\midrule
w/o Bounded Offset            & .879 & .016 & .919 & .821 & .881 & .027 & .914 & .845 \\
w/o Circular Padding          & .879 & .016 & .929 & .825 & .880 & .027 & .917 & .848 \\
w/o ON-rT2T                   & .881 & .016 & .923 & .818 & .882 & .026 & .915 & .847 \\
w/o sal-con Prior             & .879 & .016 & .922 & .817 & .880 & .029 & .909 & .831 \\
w/o foreground contrast Prior & .882 & .017 & .928 & .822 & .878 & .027 & .917 & .839 \\
w/o pano-distortion Prior     & .881 & .016 & .921 & .827 & .881 & .029 & .913 & .838 \\
w/o Gate                      & .882 & .016 & .929 & .827 & .882 & .027 & .916 & .841 \\
\midrule
\multicolumn{9}{c}{\textit{Group 3: Augmentation}} \\
\midrule
w/o Horizontal Roll & .879 & .017 & .919 & .822 & .881 & \textbf{.025} & .917 & .850 \\
w/o Color Jitter    & .883 & .015 & .929 & .826 & .879 & .027 & .915 & .838 \\
w/o Random Flip     & .884 & .015 & .928 & .827 & .883 & .026 & .913 & .846 \\
\midrule
\multicolumn{9}{c}{\textit{Group 4: Loss}} \\
\midrule
only BCE Loss  & .878 & .017 & .910 & .807 & .879 & .029 & .899 & .824 \\
w/o IoU Loss   & .885 & .016 & .922 & .809 & .881 & .028 & .901 & .825 \\
w/o Align Loss & .884 & .018 & .929 & .827 & .883 & .026 & \textbf{.919} & .850 \\
\bottomrule
\end{tabular}%
}
\end{table}

\subsection{Ablation Study}
To validate each design in SphereSOD, we conduct systematic ablation experiments on 360-SOD and ODI-SOD datasets (Table~\ref{tab_ablation}). The experiments are organized into four groups, covering the core modules, internal components, data augmentation strategies, and loss functions.

\subsubsection{Effectiveness of Modules}
We first evaluate the two core modules: PG-ME (with PG-DSA) and DPPD. As shown in Group 1, removing PG-DSA consistently degrades performance, reducing $S_m$ by 0.014 on 360-SOD and 0.016 on ODI-SOD. Replacing DPPD with a standard decoder leads to larger drops in mean$F$, confirming the importance of cascaded refinement for region accuracy and boundary quality. Removing both modules causes the most severe degradation, with clear drops in $S_m$, mean$F$, and MAE on both datasets. These results demonstrate that PG-DSA and DPPD are complementary and jointly contribute to robust panoramic SOD.

\subsubsection{Effectiveness of Components}
Group 2 analyzes the internal components. For PG-DSA, removing the bounded offset constraint or circular padding reduces $S_m$ on both datasets, indicating that constrained deformation and longitudinal continuity are both important for ERP images. In the decoder, replacing ON-rT2T with standard reverse T2T decreases mean$F$, showing that overlap normalization helps reduce upsampling bias. Removing the sal-con, foreground contrast, or pano-distortion prior also degrades performance, with the sal-con prior causing the largest mean$F$ drop on both datasets. This verifies that these priors provide complementary guidance for task-relevant context aggregation. In addition, removing the PCRD gate reduces $S_m$ and mean$F$, demonstrating its effectiveness in adaptive residual refinement.

\subsubsection{Effectiveness of Data Augmentation}
Group 3 evaluates the data augmentation strategies. The panorama-specific horizontal roll is the most effective on 360-SOD, where removing it decreases $S_m$ by 0.010 and increases MAE from 0.014 to 0.017. On ODI-SOD, it also improves $S_m$ and mean$E$, although the MAE remains competitive without it. This suggests that horizontal roll better exploits the longitudinal continuity of ERP images and improves spatial diversity. Color jitter and random flipping also bring consistent gains, providing complementary photometric and geometric regularization.

\subsubsection{Effectiveness of Loss Functions}
Group 4 studies the loss functions. Using only multi-scale BCE gives clearly inferior results on both datasets, indicating that pixel-wise supervision alone is insufficient for accurate region-level structure and global consistency. Adding IoU loss significantly improves mean$F$, especially on 360-SOD and ODI-SOD, showing its benefit for region-level overlap. Align Loss further improves the overall performance, particularly reducing MAE and improving $S_m$ on 360-SOD, while maintaining competitive results on ODI-SOD. These results confirm that BCE, IoU, and Align Loss provide complementary supervision for accurate and structurally consistent saliency prediction.

\section{Conclusion}
This paper presents SphereSOD, an ERP-native framework for panoramic SOD that couples panoramic geometry with evolving salient structures, thereby connecting geometry-adaptive feature perception with structure-conditioned progressive recovery directly in ERP space.
At the encoding stage, PG-ME initializes deformable sampling from spherical projection geometry to adapt to distortion. Meanwhile, bounded adaptive offsets constrain sampling to semantically relevant neighborhoods and circular padding preserves wrap-around continuity. 
At the decoding stage, DPPD mitigates upsampling bias via ON-rT2T, integrates foreground-contrast, saliency-contour, and distortion priors through PGTR to guide contextual aggregation, and refines boundaries via PCRD gated residual learning. In addition, the primary–auxiliary dual-branch prediction mechanism complements global semantics with local details, helping preserve structure  during progressive resolution recovery. Extensive experiments on 360-SOD, 360-SSOD, and ODI-SOD show that SphereSOD outperforms existing panoramic and 2D methods, validating the effectiveness of coupling panoramic geometry and salient structures throughout feature perception and progressive decoding.


\bibliographystyle{IEEEtran}
\bibliography{ref}

\end{document}